# Variational Approximations between Mean Field Theory and the Junction Tree Algorithm

**Wim Wiegerinck**
RWCP, Theoretical Foundation SNN,
University of Nijmegen,
Geert Grooteplein 21, 6525 EZ, Nijmegen, The Netherlands
wimw@mbfys.kun.nl

## Abstract

Recently, variational approximations such as the mean field approximation have received much interest. We extend the standard mean field method by using an approximating distribution that factorises into cluster potentials. This includes undirected graphs, directed acyclic graphs and junction trees. We derive generalised mean field equations to optimise the cluster potentials. We show that the method bridges the gap between the standard mean field approximation and the exact junction tree algorithm. In addition, we address the problem of how to choose the structure and the free parameters of the approximating distribution. From the generalised mean field equations we derive rules to simplify the approximation in advance without affecting the potential accuracy of the model class. We also show how the method fits into some other variational approximations that are currently popular.

## 1 INTRODUCTION

Graphical models, such as Bayesian networks, Markov fields, and Boltzmann machines provide a rich framework for probabilistic modelling and reasoning (Pearl, 1988; Lauritzen and Spiegelhalter, 1988; Jensen, 1996; Castillo et al., 1997; Hertz et al., 1991). Their graphical structure provides an intuitively appealing modularity and is well suited to the incorporation of prior knowledge. The invention of algorithms for exact inference during the last decades has lead to the rapid increase in popularity of graphical models in modern AI. However, exact inference is NP-hard (Cooper, 1990). This means that large, densely connected networks are intractable for exact computation, and approximations are necessary.

In this context, the variational methods gain increasingly interest (Saul et al., 1996; Jaakkola and Jordan, 1999; Jordan et al., 1999; Murphy, 1999). An advantage of these methods is that they provide bounds on the approximation error and they fit excellently into a generalised-EM framework for learning (Saul et al., 1996; Neal and Hinton, 1998; Jordan et al., 1999). This is in contrast to stochastic sampling methods (Castillo et al., 1997; Jordan, 1998) which may yield unreliable results due to finite sampling times. Until now, however, variational approximations have been less widely applied than Monte Carlo methods, arguably since their use is not so straightforward.

One of the simplest and most prominent variational approximations is the so-called mean field approximation which has its origin in statistical physics (Parisi, 1988). In the mean field approximation, the intractable distribution $P$ is approximated by a completely factorised distribution $Q$ by minimisation of the Kullback-Leibler (KL) divergence between $P$ and $Q$. Optimisation of $Q$ leads to the so-called mean field equations, which can be solved efficiently by iteration. A drawback of the standard mean field approximation is its limited accuracy due to the restricted distribution class.

For this reason, extensions of the mean field approximation have been devised by allowing the approximating distributions $Q$ to have a more rich, but still tractable structure (Saul and Jordan, 1996; Jaakkola and Jordan, 1998; Ghahramani and Jordan, 1997; Wiegerinck and Barber, 1998; Barber and Wiegerinck, 1999; Haft et al., 1999; Wiegerinck and Kappen, 2000). In this paper, we further develop this direction. In section 2 we present a general variational framework for approximate inference in an (intractable) target distribution using a (tractable) approximating distribution that factorises into overlapping cluster potentials. Generalised mean field equations are derived which are used in an iterative algorithm to optimise the cluster potentials of the approximating distribution. This procedure is guaranteed to lead to a local minimum of the KL-divergence. In section 3 we show the link between this procedure and standard exact inference methods. In section 4 we give conditions under which the complexity of the approximating model class can be reduced in advance without affecting its potential accuracy. In sections 5 and 6 we consider approximating directed graphs and we construct approximating junction trees. In section 7, we consider the approximation of target distributions for which the standard approach of KL minimisation is intractable.

# 2 VARIATIONAL FRAMEWORK

## 2.1 TARGET DISTRIBUTIONS

Our starting point is a probabilistic distribution $P(x)$ on a set of discrete variables $x = x_1, \ldots, x_n$ in a finite domain, $x_i \in \{1, \ldots, n_i\}$. Our goal is to find its marginals $P(x_i)$ on single variables or small subsets of variables $P(x_i, \ldots, x_k)$. We assume that $P$ can be written in the following factorisation

$$P(x) \;=\; \frac{1}{Z_P} \prod_\alpha \Psi_\alpha(d_\alpha) \;=\; \exp \sum_\alpha \psi_\alpha(d_\alpha) - z_P, \quad (1)$$

in which $\Psi_\alpha$ are potential functions that depend on a small number of variables, denoted by the clusters $d_\alpha$. $Z_P$ is a normalisation factor that might be unknown. Note that the potential representation is not unique. When it is convenient, we will use the logarithmic form of the potentials, $\psi_\alpha = \log \Psi_\alpha$, $z_P = \log Z_P$.

An example is a Boltzmann machine with binary units (Hertz et al., 1991),

$$P(x) = \frac{1}{Z_P} \exp(\sum_{i<j} w_{ij} x_i x_j + \sum_k h_k x_k), \quad (2)$$

that fits in our form (1) with $d_{ij} = (x_i, x_j)$, $i < j$, $d_k = x_k$ and potentials $\psi_{ij}(x_i, x_j) = w_{ij} x_i x_j$, $\psi_k(x_k) = h_k x_k$.

Another example of a distribution that fits in our framework is a Bayesian network given evidence $e$,

$$P_e(x) = P(x|e) = \frac{\prod_j P(x_j|\pi_j)}{P(e)},$$

which can be expressed in terms of the potentials $\Psi_j(d_j) = P(x_j|\pi_j)$, with $d_j = (x_j, \pi_j)$ and the normalisation $Z_P = P(e)$. This example shows that our inference problem includes the problem of computation of conditionals given evidence, since conditioning can be included by absorbing the evidence into the model definition via $P_e(x) = P(x, e)/P(e)$.

The complexity of computing marginals in $P$ depends on the underlying graphical structure of the model,

and is exponential in the maximal clique size of the triangulated moralised graph (Lauritzen and Spiegelhalter, 1988; Jensen, 1996; Castillo et al., 1997). This may lead to intractable models, even if the clusters $d_\alpha$ are small. An example is a fully connected Boltzmann machine: the clusters contain at most two variables, while the model has one clique that contains all the variables in the model.

## 2.2 APPROXIMATING DISTRIBUTIONS

In the variational method the intractable probability distribution $P(x)$ is approximated by a tractable distribution $Q(x)$. This distribution can be used to compute probabilities of interest. In the standard (mean field) approach, $Q$ is a completely factorised distribution, $Q(x) = \prod_i Q(x_i)$. We take the more general approach with $Q$ being a tractable distribution that factorises according a given structure. By tractable we mean that marginals over small subsets of variables are computationally feasible.

To construct $Q$ we first define its structure.

$$Q(x) = \frac{1}{Z_Q} \prod_\gamma \Phi_\gamma(c_\gamma) = \exp \sum_\gamma \varphi_\gamma(c_\gamma) - z_Q, \quad (3)$$

in which $c_\gamma$ are predefined clusters whose union contains all variables. $\Phi_\gamma(c_\gamma)$ are nonnegative potentials of the variables in the clusters. The only restriction on the potentials is the global normalisation

$$\sum_{\{x\}} \prod_\gamma \Phi_\gamma(c_\gamma) \;=\; Z_Q. \quad (4)$$

## 2.3 VARIATIONAL OPTIMISATION

The approximation $Q$ is optimised such that the Kullback-Leibler (KL) divergence between $Q$ and $P$,

$$D(Q, P) = \sum_{\{x\}} Q(x) \log \frac{Q(x)}{P(x)} \equiv \left\langle \log \frac{Q(x)}{P(x)} \right\rangle,$$

is minimised. In this paper, $\langle \ldots \rangle$ denotes the average with respect to $Q$. The KL-divergence is related to the difference of the probabilities of $Q$ and $P$,

$$\max_A |P(A) - Q(A)| \le \sqrt{\frac{1}{2} D(Q, P)},$$

for any event $A$ in the sample space (see (Whittaker, 1990)). In the logarithmic potential representations of $P$ and $Q$, the KL-divergence is

$$D(Q, P) \;=\; \left\langle \sum_\gamma \varphi_\gamma(c_\gamma) - \sum_\alpha \psi_\alpha(d_\alpha) \right\rangle$$
$$+ \text{ constant},$$

which shows that $D(Q, P)$ is tractable (up to a constant) when $Q$ is tractable and the clusters in $P$ and $Q$ are small.

To optimise $Q$ under the normalisation constraint (4), we do a constrained optimisation of the KL-divergence with respect to $\varphi_\gamma$ using Lagrangian multipliers. In this optimisation, the other potentials $\varphi_\beta$, $\beta \neq \gamma$ remain fixed. This leads to the solution $\varphi_\gamma^*(c_\gamma)$, given by the generalised mean field equations

$$\varphi_\gamma^*(c_\gamma) = \left\langle \sum_{\alpha \in D_\gamma} \psi_\alpha(d_\alpha) - \sum_{\beta \in C_\gamma} \varphi_\beta(c_\beta) \right\rangle_{c_\gamma} - z. \quad (5)$$

The average $\langle \ldots \rangle_{c_\gamma}$ is taken with respect to the conditional distribution $Q(x|c_\gamma)$. In (5), $D_\gamma$ (resp. $C_\gamma$) are the sets of clusters $\alpha$ in $P$ (resp. $\beta \neq \gamma$ in $Q$) that depend on $c_\gamma$. In other words, $\alpha \notin D_\gamma$ implies $Q(d_\alpha|c_\gamma) = Q(d_\alpha)$, etc. Finally, $z$ is a constant that can be inferred from the normalisation (4), i.e.

$$z = \log \sum_{\{x\}} \exp \left[ \sum_{\beta \neq \gamma} \varphi_\beta(c_\beta) + \right.$$
$$\left. + \left\langle \sum_{\alpha \in D_\gamma} \psi_\alpha(d_\alpha) - \sum_{\beta \in C_\gamma} \varphi_\beta(c_\beta) \right\rangle_{c_\gamma} \right] - z_Q . \quad (6)$$

Since $Q(x|c_\gamma)$ is independent of the potential $\varphi_\gamma$, $z$ (6) is independent of $\varphi_\gamma$. Consequently, the right hand side of (5) is independent of $\varphi_\gamma$ as well. So (5) provides a unique solution $\varphi_\gamma^*$ to the optimisation of the potential of cluster $\gamma$. This solutions corresponds to the global minimum of $D(Q, P)$ given that the potentials of other clusters $\beta \neq \gamma$ are fixed. This means that in a sequence where at each step different potentials are selected and updated, the KL-divergence decreases at each step. Since $D(Q, P) \geq 0$, we conclude that this iteration over all clusters of variational potentials leads to a local minimum of $D(Q, P)$.

In the mean field equations (5), the constant $z$ plays only a minor role and can be set to zero if desired. This can be achieved by simultanously shifting $z_Q$ and $\varphi_{c_\gamma}$ by the same amount before we optimize $\varphi_{c_\gamma}$. (This shift does not affect $Q$).

The generalized mean field equations (5) straightforwardly generalizes upon the standard mean field equations for fully factorized approximations (see e.g. (Haft et al., 1999)). The main difference is that the contribution of the other potentials $\phi_\beta$, $\beta \in C_\gamma$ vanishes in the fully factorized approximation.

In figure 1, a simple example is given.

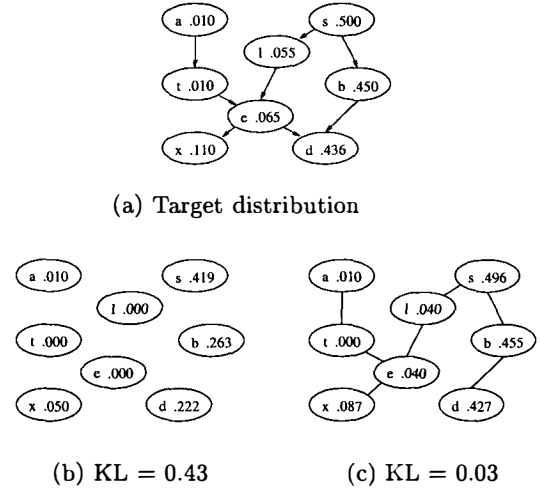

(a) Target distribution

(b) KL = 0.43       (c) KL = 0.03

Figure 1: Chest clinic model (ASIA), from (Lauritzen and Spiegelhalter, 1988). (a): Exact distribution $P$ with marginal probabilities. (b-c): Approximating distributions with approximated marginal probabilities. In (b) $Q$ is fully factorised. In (c), $Q$ is a tree. KL is the KL-divergence $D(Q, P)$ between the approximating distribution $Q$ and the target distribution $P$.

## 3 GLOBAL CONVERGENCE

In this section, we link the mean field approximation with exact computation by showing global convergence for approximations $Q$ that satisfy the following two conditions: (1) Each cluster $d_\alpha$ of $P$ is at least contained in one of the clusters $c_\gamma$ in $Q$. (2) $Q$ satisfies the so-called running intersection property.

For a definition of the running intersection property, we follow (Castillo et al., 1997): $Q$ satisfies the running intersection property if there is an ordering of the clusters of $Q$, $(c_1, \ldots, c_m)$ such that $s_\gamma \equiv c_\gamma \cap (c_1 \cup \ldots \cup c_{\gamma-1})$ is contained in at least one of the clusters $(c_1, \ldots, c_{\gamma-1})$.

If a cluster $c_\gamma$ intersects with the separator $s_\delta$ of a successor $c_\delta$, there are three possibilities: $s_\delta$ is contained in another successor $c_\eta$ $(\delta > \eta > \gamma)$, or $s_\delta$ is contained in $c_\gamma$ itself, or $s_\delta$ intersects only with the separator $s_\gamma$ (since $s_\delta$ is contained in a predecessor of $c_\gamma$, which is separated by $s_\gamma$). We denote $\Delta_\gamma = \{s_\delta \subset c_\gamma | s_\delta \not\subset c_\eta, \delta > \eta > \gamma\}$. So each separator is contained in exactly one $\Delta_\gamma$. Finally, we define $A_\gamma = \{d_\alpha \subset c_\gamma | d_\alpha \not\subset c_\eta, \eta > \gamma\}$. Each cluster of $P$ is contained in exactly one $A_\gamma$.

With these preliminaries, we consider the mean field equations (5) applied to the potentials of $Q$. We con-sider a decreasing sequence of updates. At first, the last potential $\phi_m$ is updated. This results in

$$\phi_m^\star(c_m) = \sum_{\alpha \in A_m} \psi_\alpha(d_\alpha) + \tilde{\phi}_m(s_m) \,,$$

where $\tilde{\phi}_m(s_m)$ is a function that depends only on the value of the separator $s_m$. If $s_m$ is empty, $\tilde{\phi}_m(s_m)$ is a constant. If in this sequence, potential $\phi_\gamma$ has its turn, the result is

$$\phi_\gamma^\star(c_\gamma) = \sum_{\alpha \in A_\gamma} \psi_\alpha(d_\alpha) - \sum_{\delta \in \Delta_\gamma} \tilde{\phi}_\delta(s_\delta) + \tilde{\phi}_\gamma(s_\gamma), \quad (7)$$

where, again, $\tilde{\phi}_\gamma(s_\gamma)$ is a function that depends only on the value of the separator $s_\gamma$. Finally, after all potentials have been updated, we add up all potentials and obtain

$$\sum_\gamma \phi_\gamma^\star(c_\gamma) = \sum_\alpha \psi_\alpha(d_\alpha) + z \,,$$

which shows that $Q$ converged to $P$ in one sweep of updates. If the sequence of updates is in random order, the result shows convergence in finite time. Note that if condition 2 – the running intersection property – is not satisfied, the mean field procedure does not need to converge to the global optimum, even if the model class of $Q$ is rich enough to model $P$ exactly (condition 1).

Standard exact inference methods (Lauritzen and Spiegelhalter, 1988; Jensen, 1996; Castillo et al., 1997), (after constructing cluster-sets that satisfy the two above-stated conditions), are very similar to (7). The difference is that standard exact methods just keep the separator functions $\tilde{\phi}_\gamma(s_\gamma)$ equal to zero (which is of course much more efficient). The advantage of the generalised mean field approximation is that it generalises to $Q$'s that do not meet the required conditions for exact computation.

## 4 EXPLOITING SUBSTRUCTURES

An obviously important question is how to choose the structure of $Q$ to get the best compromise between approximation error and complexity. Another question is if our approach, in which all the potentials of $Q$ are fully adaptive, is the best way to go. An alternative approach, originally proposed in (Saul and Jordan, 1996), is to copy the target distribution $P$ and remove potentials that makes $P$ intractable. The removed potentials are compensated by introducing additional variational parameters in the remaining potentials. In the context of our paper, this can be expressed as: pick a subset $A$ of clusters $\alpha$ of the distribution $P$, copy the

potentials of $A$ and parametrise the approximation $Q$ as

$$Q(x) = \exp\left[\sum_\gamma \phi_\gamma(c_\gamma) + \sum_{\alpha \in A} \psi_\alpha(d_\alpha) - z_Q\right]. \quad (8)$$

The (variational) potentials $\phi_\gamma$ are to be optimised. The potentials $\psi_\alpha$, $\alpha \in A$ are copies of the potentials in the target distribution, and are fixed. The clusters $c_\gamma$ and $d_\alpha$, $\alpha \in A$ define the cluster-set of $Q$, and they contain all variables. The approximation (8) is of the general form as (3). The difference is that in (8) some potentials are set in advance to specific values, and do not need to be optimised any more. This has obviously big computational advantages. A disadvantage is that the copied potentials might be suboptimal, and that by fixing these potentials the method might be weaker than one in which they are adaptive.

From the mean field equations (5), one can infer conditions under which the optimisation effectively uses copied potentials of $P$, and simplifies free parameters of $Q$, and thus effectively restricts the model class. This is stated in the following.

*Lemma* Let $Q$ be parametrised as in (8) and let $c_\kappa$ be one of the clusters of $Q$. If $c_\kappa$ can be written as a union

$$c_\kappa = \bigcup_{\alpha \in A_\kappa} d_\alpha \cup \bigcup_u c_{\kappa^u},$$

with $u = 1, \ldots, u_{\max}$, (nb., the $c_{\kappa^u}$'s are not in the cluster-set of $Q$), such that for all of the remaining clusters $t$ in $P$ and $Q$, i.e., $t \in \{d_\alpha, c_\gamma | \alpha \notin A \cup A_\kappa, \gamma \neq \kappa\}$, the independency

$$Q(t|c_\kappa) = Q(t|c_{\kappa^u})$$

holds for at least one $u \in \{1, \ldots, u_{\max}\}$, regardless of the values of the potentials $\phi$ and $\psi$, then the optimised approximating distribution $Q$ (8) takes the form

$$Q(x) = \exp\left[\sum_u \phi_{\kappa^u}(c_{\kappa^u}) + \sum_{\gamma \neq \kappa} \phi_\gamma(c_\gamma) + \right.$$
$$\left. + \sum_{\alpha \in A \cup A_\kappa} \psi_\alpha(d_\alpha) - z_q\right].$$

This is straightforwardly verified by applying the mean field optimisation to $\phi_\kappa$ in $Q$.

From this lemma, considerable simplifications can be deduced. Consider, for example, a fully connected Boltzmann machine $P$ (cf (2)) approximated by $Q$. If $Q$ consists of potentials of non-overlapping clusters $c_\gamma$, it can inferred that the optimised $Q$ will consists of the fixed copies of the weights of $P$ that are within the clusters $c_\gamma$ of $Q$, adaptive biases for the nodes that are connected with weights in $P$ which are not copied into $Q$, and fixed copies of biases for the remaining nodes.

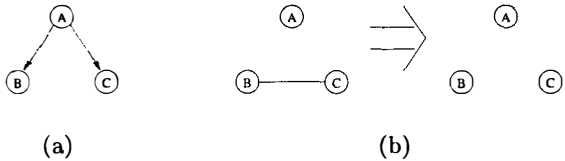

(a)                                    (b)

Figure 2: Example of redundant structure. (a): Graph of exact distribution $P(A)P(B|A)P(C|A)$. (b): Optimisation of an approximating distribution with structure $Q(A)Q(B,C)$ leads to a distribution with simpler structure $Q(A)Q(B)Q(C)$. The variables $B$ and $C$ become independent in $Q$, although they are marginally dependent in $P$ (via $A$).

Note that optimal weights in an approximation of a Boltzmann machine are not always just copies from the target distribution. An illustrative counter example is the target distribution $P(x_1, x_2, x_3) \propto \exp(w_{12}x_1x_2 + w_{13}x_1x_3 + w_{23}x_2x_3)$ with $w_{12} = \infty$, so $x_1$ and $x_2$ are hard coupled ($x_i = \pm 1$). The optimal approximation of the form $Q(x_1, x_2, x_3) \propto \Phi(x_1, x_2)\Phi(x_2, x_3)$ is given by $\Phi(x_1, x_2) \propto \exp(w_{12}x_1x_2)$ and $\Phi(x_2, x_3) \propto \exp([w_{13} + w_{23}]x_2x_3)$. The approximation in which the weight between $x_2$ and $x_3$ in $Q$ is copied from $P$ (i.e. $w_{23}$ instead of $w_{13} + w_{23}$ ) is suboptimal.

The convergence times between approximate models with and without using copied potentials may differ, even if their potential accuracies are the same. As an example, consider the target $P(x_1, x_2) \propto \exp(w_{12}x_1x_2)$. The approximation $Q(x_1, x_2) \propto \Phi(x_1, x_2)$ convergences in one step. On the other hand, in $Q(x_1, x_2) \propto \exp(w_{12}x_1x_2 + \phi_1(x_1) + \phi_2(x_2))$, the potentials $\phi_i$ decay only exponentially.

The lemma generalises the result on graph partitioning in Boltzmann machines as presented in (Barber and Wiegerinck, 1999). It shows and clarifies in which cases the copied potentials of tractable substructures as originally proposed in (Saul and Jordan, 1996) are optimal. A nice example in which the copied potentials are optimal is the application to the Factorial Hidden Markov Models in (Ghahramani and Jordan, 1997; Jordan et al., 1999).

The lemma provides a basis to the intuition that adding structure to $Q$ that is not present in $P$ might be redundant. (The lemma is still valid if $A_\kappa$ is empty.) In fig. 2 a simple example is given. A similar result for approximations using directed graphs (cf. section 5) is obtained in (Wiegerinck and Kappen, 2000). Finally, we note that the lemma only provides sufficient conditions for simplification.

## 5   DIRECTED APPROXIMATIONS

A slightly different class of approximated distributions are the 'directed' factorisations. These have been considered previously in (Wiegerinck and Barber, 1998; Barber and Wiegerinck, 1999; Wiegerinck and Kappen, 2000), but they fit well in the more general framework of this paper. Directed factorisations can be written in the same form (3), but the clusters need to have an ordering $c_1, c_2, c_3, \ldots$. We define separator sets $s_\gamma = c_\gamma \cap \{c_1 \cup \ldots \cup c_{\gamma-1}\}$ and residual sets $r_\gamma = c_\gamma \backslash s_\gamma$. We restrict the potentials $\Phi_\gamma(c_\gamma) = \Phi_\gamma(r_\gamma, s_\gamma)$ to satisfy the local normalisation

$$\sum_{\{r_\gamma\}} \Phi_\gamma(r_\gamma, s_\gamma) = 1 \,, \qquad (9)$$

We can identify $\Phi_\gamma(r_\gamma, s_\gamma) = Q(r_\gamma|s_\gamma)$ and (3) can be written in the familiar directed notation

$$Q(x) = \prod_\gamma Q(r_\gamma|s_\gamma) \,.$$

To optimise the potentials $\varphi_\gamma(r_\gamma, s_\gamma)$ ($= \log Q(r_\gamma|s_\gamma)$), we do again a constraint optimisation with constraints (9). This leads to generalised mean field equations for directed distributions

$$\varphi_\gamma^*(r_\gamma, s_\gamma) = \left\langle \sum_{\alpha \in D_\gamma^r} \psi_\alpha(d_\alpha) - \sum_{\beta \in C_\gamma^r} \varphi_\beta(c_\beta) \right\rangle_{r_\gamma, s_\gamma} - z(s_\gamma),$$

in which $D_\gamma^r$ (resp. $C_\gamma^r$) is the set of clusters $\alpha$ in $P$ (resp. $\beta \neq \gamma$ in $Q$) that depend on $r_\gamma$. $z(s_\gamma)$ is a local normalisation factor that can be inferred from (9), i.e.

$$z(s_\gamma) = \log \sum_{\{r_\gamma\}} \exp \left\langle \sum_{\alpha \in D_\gamma} \psi_\alpha(d_\alpha) - \sum_{\beta \in C_\gamma} \varphi_\beta(c_\beta) \right\rangle_{c_\gamma} \,.$$

## 6   JUNCTION TREES

For the definition of junction trees, we follow (Jensen, 1996): A cluster tree is a tree of clusters of variables which are linked via separators. These consists of the variables in the adjacent clusters. A cluster tree is a junction tree if for each pair of clusters $c_\gamma, c_\delta$, all nodes in the path between $c_\gamma$ and $c_\delta$ contain the intersection. In a consistent junction tree, the potentials $\Phi_\gamma$ and $\Phi_\delta$ of the nodes $c_\gamma$, $c_\delta$ with intersection $I$ satisfy

$$\sum_{c_\gamma \backslash I} \Phi_\gamma(c_\gamma) = \sum_{c_\delta \backslash I} \Phi_\delta(c_\delta) \,.$$

We consider consistent junction tree representations of $Q$ of the form

$$Q(x) = \frac{\prod_\gamma \Phi_\gamma(c_\gamma)}{\prod_{(\gamma, \delta)} \Phi_{\gamma, \delta}(s_{\gamma, \delta})} \,,$$

in which the product in the denominator is taken over the separators. The separator potentials are defined by the cluster potentials,

$$\Phi_{\gamma,\delta}(s_{\gamma,\delta}) = \sum_{c_\gamma \backslash s_{\gamma,\delta}} \Phi_\gamma(c_\gamma) \ .$$

The junction tree representation is convenient, because the cluster probabilities can directly be read from the cluster potentials:

$$Q(c_\gamma) = \Phi_\gamma(c_\gamma) \ .$$

For a more detailed treatment of junction trees we refer to (Jensen, 1996).

In the following, we show how approximations can be optimised while maintaining the junction tree representation. Taking one of the clusters, $c_\kappa$, separately, we write $Q$ as the potential $\Phi_\kappa$ times $Q(x|c_k)$

$$Q(x) = \Phi_\kappa(c_\kappa) \times \frac{\prod_{\gamma \neq \kappa} \Phi_\gamma(c_\gamma)}{\prod_{(\gamma,\delta)} \Phi_{\gamma,\delta}(s_{\gamma,\delta})}$$

Subsequently, we update $\Phi_\kappa$ according to the mean field equations (5),

$$\Phi_\kappa^*(c_\kappa) = \frac{1}{Z} \exp \left\langle \sum_{d_\alpha \in D_\kappa} \log \Psi(d_\alpha) \right.$$
$$\left. - \sum_{\gamma \in C_\kappa} \log \Phi_\gamma(c_\gamma) + \sum_{(\gamma,\delta) \in S_\kappa} \log \Phi_{\gamma,\delta}(s_{\gamma,\delta}) \right\rangle_{c_\kappa}, \quad (10)$$

where $S_\kappa$ is the set of separators that depend on $c_\kappa$. $Z$ makes sure that $\Phi_\kappa^*$ is properly normalised. Now, however, the junction tree is not consistent anymore. We can fix this by applying the standard *DistributeEvidence*$(c_\kappa)$ operation to the junction tree (see (Jensen, 1996)). In this routine, $c_\kappa$ sends a message to all its neighbours $c_\gamma$ via

$$\Phi_{\gamma,\kappa}^*(s_{\gamma\kappa}) = \sum_{c_\kappa \backslash s_{\gamma\kappa}} \Phi_\kappa^*(c_\kappa)$$

and

$$\Phi_\gamma^*(c_\gamma) = \Phi_\gamma(c_\gamma) \frac{\Phi_{\gamma,\kappa}^*(s_{\gamma\kappa})}{\Phi_{\gamma,\kappa}(s_{\gamma\kappa})} \ .$$

Recursively, the neighbours $c_\gamma$ send messages to all their neighbours except the one from which the message came. After this procedure, the junction tree is consistent again, and another potential can be updated by (10).

Since the *DistributeEvidence* routine does not change the distribution $Q$ (it only makes it consistent), the global convergence result (section 3) applies if the structure of $Q$ is a junction tree of $P$. This links the mean field theory with the exact junction tree algorithm.

# 7 APPROXIMATED MINIMISATION

The complexity of the variational method is at least proportional to the number of states in the clusters $d_\alpha$ of the target distribution $P$, since it requires the computation of averages of the form $\langle \psi(d_\alpha) \rangle$. In other words, the method presented in this paper can only be computationally tractable if the number of states in $d_\alpha$ is reasonably small. If the cluster potentials are explicitly tabulated, the required storage space is also proportional with the number of possible cluster states. In practice, potentials with large number of cluster states are parameterised. In these cases, one can try to exploit the parametrisation and approximate $\langle \psi(d_\alpha) \rangle$ by a tractable quantity.

Examples are target distributions $P$ with conditional probabilities $P(x_i|\pi_i)$ that are modelled as noisy-OR gates (Pearl, 1988; Jensen, 1996) or as weighted sigmoid functions (Neal, 1992). For these parametrisations $\langle \log P(x_i|\pi_i^P) \rangle$ can be approximated by a tractable quantity $\mathcal{E}_i(Q, \xi)$ (which may be defined using additional variational parameters $\xi$). As an example, consider tables parametrised as sigmoid functions,

$$P(x_i = 1|\{x_k\}) = \sigma(z_i) \equiv (1 + \exp(z_i))^{-1} \ ,$$

where $z_i$ is the weighted input of the node, $z_i = \sum_k w_{ik} x_k + h_i$. In this case, the averaged log probability is intractable for large parent sets. To proceed we can use the approximation proposed in (Saul et al., 1996)

$$\langle \log(1 + e^{z_i}) \rangle \leq$$
$$\xi_i \langle z_i \rangle + \log \left\langle e^{-\xi_i z_i} + e^{(1-\xi_i)z_i} \right\rangle \equiv \mathcal{E}_i(Q, \xi) \ ,$$

which is tractable if $Q$ is tractable (Wiegerinck and Barber, 1998; Barber and Wiegerinck, 1999). Numerical optimisation of $\mathcal{L}(Q, \xi) \equiv \langle \log Q \rangle - \mathcal{E}(Q, \xi)$ with respect to $Q$ and $\xi$ leads to local minimum of an upper bound of the KL-divergence. Note however, that iteration of fixed point equations derived from $\mathcal{L}(Q, \xi)$ does not necessarily lead to convergence, due to the nonlinearity of $\mathcal{E}$ with respect to $Q$. In (Wiegerinck and Kappen, 2000) numerical simulations are performed on artificial target distributions $P$ that had tractable substructures as well as sigmoidal nodes with large parent sets. Target distributions with varying system size were approximated by fully factorised distributions as well as distributions with structure. The results showed that an approximation using structure can improve significantly the accuracy of approximation within feasible computer time. This seemed independent of the problem size.

Another example is a hybrid Bayesian network (which has continuous and discrete variables). In the remainder of this section we closely follow (Murphy, 1999). For expositional clarity we consider the distribution $P(r|x)P(x|t)P(t)$, in which $r$ and $t$ are binary variables and $x$ is a continuous variable. The conditional distribution $P(r|x)$ is parametrised by a sigmoid, $P(r = 1|x) = \sigma(wx + b)$ with parameters $w$ and $b$. The conditional distribution $P(x|t)$ is a conditional Gaussian, $P(x|t) = \exp(g_t + xh_t + xK_t/2)$ in which $(h_t, K_t)$ are parameters depending on $t$ and $P(t)$ is a simple table with two entries. As (Murphy, 1999) showed, computation of the conditional distributions of $x$ and $t$ given observation of $r$ is difficult. In (Murphy, 1999), it is proposed to approximate the KL-divergence by using the quadratic lower bound of the sigmoid function (Jaakkola, 1997; Murphy, 1999),

$$\log \sigma(x) \geq x/2 + \lambda(\xi)x^2 + \log \sigma(\xi) - \xi/2 - \xi^2\lambda(\xi),$$

with $\lambda(\xi) = -1/4\xi \tanh(\xi/2)$. By fixing $\xi$, this bound leads to an tractable upper bound of the KL-divergence

$$\mathcal{L}(Q, \xi) = \langle \log Q(t) + \log Q(x|t) - \log P(t)$$
$$-(g_t + g(\xi) + x(h_t + h(\xi)) + x^2(K_t + K(\xi))/2) \rangle,$$

in which

$$g(\xi) = \log \sigma(\xi) + \tfrac{1}{2}(2r - 1)b - \tfrac{1}{2}\xi + \lambda(\xi)(b^2 - \xi^2)$$
$$h(\xi) = \tfrac{1}{2}(2r - 1)w + 2\lambda(\xi)bw$$
$$K(\xi) = 2\lambda(\xi)w^2$$

For given $\xi$, the optimal distribution $Q(x, t)$ is simply given by

$$Q(x, t) \propto P(t) \exp\left(g_t + g(\xi) + x(h_t + h(\xi)) + x^2(K_t + K(\xi))/2\right).$$

In other words, $Q$ is the product of a conditional Gaussian and a table. Since the parameters of the conditional Gaussian $Q(x|t)$ depends on $t$, an obvious extension of this scheme is to make the contribution that depends on $\xi$ also depending on $t$. In other words, we replace the single parameter $\xi$ by two parameters $\xi_t$, $t = 0, 1$, and bound the KL-divergence with

$$\mathcal{L}(Q, \xi_t) = \langle \log Q(t) + \log Q(x|t) - \log P(t)$$
$$-(g_t + g(\xi_t) + x(h_t + h(\xi_t)) + x^2(K_t + K(\xi_t))/2) \rangle.$$

Then it follows that for given $\xi_t$, the optimal distribution $Q(x, t)$ is given by

$$Q(x, t) \propto P(t) \exp\left(g_t + g(\xi_t) + x(h_t + h(\xi_t)) + x^2(K_t + K(\xi_t))/2\right).$$

To optimise $\xi_t$, we find in analogy with (Murphy, 1999)

$$\xi_t^2 = \left\langle (wx + b)^2 \right\rangle_t.$$

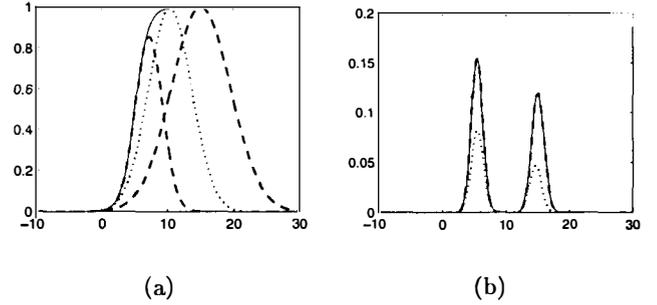

(a)                              (b)

Figure 3: The effect of using approximations with structure in hybrid networks. We show results for a network $P(t)P(x|t)P(r|x)$, in which $t$ and $r$ are binary (0/1) and $x$ is continuous. $P(t = 1) = 0.3$, $P(x|t)$ is a conditional Gaussian with $\mu_{0,1} = (10, 20)$, $\sigma_{0,1} = 1$ and $p(r|x)$ is defined using a sigmoid with $w = -1$, $b = 5$. (This example is based on the crop network, with $t$ is 'subsidy', $x$ is 'price' and $r$ is 'buy'. 'Crop' is assumed to be observed in its mean value - see (Murphy, 1999) for details). In (a) we plot $\sigma(-(wx - b))$ as a function of $x$ (solid), as well as the variational lower bound using one optimised unconditional parameter $\xi$ (dotted) and the two bounds for the optimised conditional variational parameters $\xi_t$ (dashed). In (b) we plot $P(r = 0, x)$ as a function of $x$ using the exact probability (solid), the approximation using one unconditional parameter $\xi$ (dotted) and two conditional parameters $\xi_t$ (dashed - this graph coincides with the exact graph).

The largest improvements of this extension can be expected when the posterior distribution (given observation of $r$) is multi-modal. In figure 3, an example is given.

## 8    DISCUSSION

Finding accurate approximations of graphical models such as Bayesian networks is crucial if their application to large scale problems is to be realised. We have presented a general scheme to use a (simpler) approximating distribution that factorises according to a given structure. The scheme includes approximations using undirected graphs, directed acyclic graphs and junction trees. The approximating distribution is tuned by minimisation of the Kullback-Leibler divergence. We have shown that the method bridges the gap between standard mean field theory and exact computation. We have contributed to a solution for the question how to select the structure of the approximating distribution, and when potentials of the target distribution can be exploited. Parametrised dis-tributions with large parent sets can be dealt with by minimising an approximation of the KL-divergence. In the context of hybrid networks, we showed that it can be worthwhile to endow possible additional variational parameters with structure as well.

An open question is still how to find good and efficient structures for the variational quantities. Nevertheless, one of conclusions of this paper is that using (more) structure in variational quantities is worthwhile to try if increase in accuracy is needed. It is often compatible within the used variational method (EM-learning, applications to hybrid networks etc), and is often possible without too much computational overhead. As such, variational methods provide a flexible tool for approximation in which accuracy and efficiency can be tuned to the needs and the computational resources of the application.

### Acknowledgements

I would like to thank David Barber, Ali Taylan Cemgil, Tom Heskes and Bert Kappen for useful discussions. I thank Ali Taylan Cemgil for sharing his matlab routines.